\pdfoutput=1

\documentclass[11pt]{article}

\usepackage{acl}

\usepackage{times}
\usepackage{latexsym}

\usepackage[T1]{fontenc}

\usepackage[utf8]{inputenc}

\usepackage{microtype}
\usepackage{xspace,mfirstuc,tabulary}
\usepackage{enumitem}
\usepackage{amsthm}
\usepackage{amssymb}
\usepackage{amsmath}
\usepackage{multirow}
\usepackage{comment}
\usepackage{array}
\usepackage{tabularx}
\definecolor{darkspringgreen}{rgb}{0.09, 0.45, 0.27}
\usepackage{algorithm,algpseudocode}
\newlength{\bibitemsep}
\newlength{\bibparskip}
\let\oldthebibliography\thebibliography
\renewcommand\thebibliography[1]{%
  \oldthebibliography{#1}%
  \setlength{\parskip}{\bibitemsep}%
  \setlength{\itemsep}{\bibparskip}%
}

\usepackage{enumitem}
\setlist{nolistsep}
\theoremstyle{definition}
\newtheorem{definition}{Definition}[section]

\newcommand{\R}{\mathbb{R}}
\newcommand{\D}{\mathbb{D}}
\newcolumntype{P}[1]{>{\centering\arraybackslash}p{#1}}

\usepackage{booktabs}
\usepackage{subcaption}
\usepackage[justification=centering]{caption}

\usepackage{supertabular}

\title{Detecting Textual Adversarial Examples Based on \\ Distributional Characteristics of Data Representations}

\author{Na Liu$^1$ Mark Dras$^1$ Wei Emma Zhang$^2$\\
$^1$ School of Computing, Macquarie University \\
$^2$ School of Computer Science, The University of Adelaide \\
\texttt{na.liu8@students.mq.edu.au,mark.dras@mq.edu.au}\\
\texttt{wei.e.zhang@adelaide.edu.au}}

\begin{document}
\maketitle
\begin{abstract}
Although deep neural networks have achieved state-of-the-art performance in various machine learning tasks, adversarial examples, constructed by adding small non-random perturbations to correctly classified inputs, successfully fool highly expressive deep classifiers into incorrect predictions.
Approaches to adversarial attacks in natural language tasks have boomed in the last five years using character-level, word-level, phrase-level, or sentence-level textual perturbations.
While there is some work in NLP on defending against such attacks through proactive methods, like adversarial training, there is to our knowledge no effective general reactive approaches to defence via detection of textual adversarial examples such as is found in the image processing literature.
In this paper, we propose two new reactive methods for NLP to fill this gap, which unlike the few limited application baselines from NLP are based entirely on distribution characteristics of learned representations: we adapt one from the image processing literature (Local Intrinsic Dimensionality (LID)), and propose a novel one (MultiDistance Representation Ensemble Method (MDRE)).
Adapted LID and MDRE obtain state-of-the-art results on character-level, word-level, and phrase-level attacks on the IMDB dataset as well as on the later two with respect to the MultiNLI dataset.
For future research, we publish our code \footnote{Code available at \url{https://github.com/NaLiuAnna/MDRE}}.
\end{abstract}

\section{Introduction}
\label{sec:Introduction}
Highly expressive deep neural networks are fragile against adversarial examples, constructed by carefully designed small perturbations of normal examples, that can fool deep classifiers to make wrong predictions \citep{szegedy2013intriguing}.
Crafting adversarial examples in images involves adding small non-random perturbations to many pixels in inputs that should be correctly classified by a target model.  
These perturbations can force high-efficacy models into incorrect classifications and are often imperceptible to humans \citep{szegedy2013intriguing, goodfellow2014explaining, moosavi2016deepfool, papernot2016limitations, carlini2017towards, chen2018ead}.
However, when adversarial examples have been studied in the context of text, to our knowledge, only \citet{miyato2016adversarial} aligns closely with the original intuition of adversarial examples in applying perturbations to word embeddings, which are inputs of deep neural nets.
Rather, most adversarial attack techniques use more practical semantics-preserving textual changes other than embedding perturbations, at character-level, word-level, phrase-level, or sentence-level \citep{pruthi-etal-2019-combating, jia-liang-2017-adversarial, alzantot-etal-2018-generating, ribeiro-etal-2018-semantically,
ren-etal-2019-generating, iyyer-etal-2018-adversarial,  yoo-qi-2021-towards-improving, li-etal-2020-bert-attack, li-etal-2021-contextualized, jin2020bert}; see Table~\ref{tab:text_adv}. 
This variety increases the difficulty of detecting textual adversarial examples.

\begin{table*}[t]
\centering
\small
\begin{tabular}{>{\centering\arraybackslash}m{0.18\linewidth}m{0.64\linewidth}m{0.08\linewidth}}
& \textbf{Example} \centering &\textbf{Prediction} \\
\hline
\textbf{Original} & \textbf{\textcolor{darkspringgreen}{This}} is a \textbf{\textcolor{darkspringgreen}{story}} of two \textbf{\textcolor{darkspringgreen}{misfits who}} don't \textbf{\textcolor{darkspringgreen}{stand}} a \textbf{\textcolor{darkspringgreen}{chance}} \textbf{\textcolor{darkspringgreen}{alone}}, but \textbf{\textcolor{darkspringgreen}{together}} they are \textbf{\textcolor{darkspringgreen}{magnificent}}.  &  Positive\\ \hline
\textbf{Character-level}     \citep{pruthi-etal-2019-combating} & \textbf{\textcolor{orange}{TZyTis}} is a \textbf{\textcolor{orange}{sotry}} of two \textbf{\textcolor{orange}{misifts}} who don't \textbf{\textcolor{orange}{stad}} a \textbf{\textcolor{orange}{ccange}} \textbf{\textcolor{orange}{alUone}}, but \textbf{\textcolor{orange}{tpgthr}} they are \textbf{\textcolor{orange}{mgnificent}}.  & Negative\\ \hline
\textbf{Word-level} \quad    \citep{alzantot-etal-2018-generating} & This is a \textbf{\textcolor{orange}{conte}} of two \textbf{\textcolor{orange}{who}} don't \textbf{\textcolor{orange}{stands}} a \textbf{\textcolor{orange}{opportunities}} alone, but together they are \textbf{\textcolor{orange}{opulent}}. & Negative\\ \hline
\textbf{Phrase-level} \qquad \citep{iyyer-etal-2018-adversarial} & Why don't you have two misfits who don't stand a chance alone, but together they're beautiful.  & Negative\\ \hline
\textbf{Sentence-level} \qquad
\citep{jia-liang-2017-adversarial} & This is a story of two misfits who don't stand a chance alone, but together they are magnificent. \textbf{\textcolor{orange}{ready south hundred at size expected worked whose turn poor.}}  & Negative\\ \hline
\end{tabular}
\caption{\small {Examples of textual adversarial instances on a sentiment analysis task}}
\label{tab:text_adv}
\vspace{-10pt}
\end{table*}

Generating adversarial examples to attack deep neural nets and protecting deep neural nets from adversarial examples have been extensively studied in image classification tasks \citep{szegedy2013intriguing, goodfellow2014explaining, moosavi2016deepfool, papernot2016limitations,  carlini2017towards, chen2018ead, papernot2016distillation, feinman2017detecting, ma2018characterizing, lee2018simple}.
However, in the natural language domain, only crafting of adversarial examples has been comprehensively considered \citep{jia-liang-2017-adversarial, alzantot-etal-2018-generating, ribeiro-etal-2018-semantically,
ren-etal-2019-generating, iyyer-etal-2018-adversarial}.
Defence against textual adversaries, primarily through increasing the robustness of deep neural networks, is much less studied \citep{jia-etal-2019-certified,pruthi-etal-2019-combating}. 
In the image processing space, \citet{CVPRPaper} refers to these as \textit{proactive} defence methods, and \citet{carlini2017adversarial} notes that they can be evaded by optimization-based attacks, such as constructing new loss functions; in the NLP space, \citet{yoo-qi-2021-towards-improving} observes that generating word-level textual adversaries for proactive adversarial training are computationally expensive because of necessary search and constraints based on sentence encoding. 
Consequently, \citet{feinman2017detecting, ma2018characterizing, lee2018simple, papernot2018deep} explore \textit{reactive} defence methods \citep{CVPRPaper} in the image processing space: these focus on distinguishing real from adversarial examples, in order to detect them before they are passed to neural networks.  
These reactive defences have been explored in only a limited way in the NLP space \citep{mozes-etal-2021-frequency}.  Importantly, these few methods rely on procedures like testing word substitutions, quite unlike those in the image processing space, which are functions of the learned representations.

The contributions of this paper are two textual adversarial reactive detectors as follows:
\begin{itemize}
    \item Adapting the Local Intrinsic Dimensionality (LID) method from image processing to the text domain.
    \item Proposing a MultiDistance Representation Ensemble Method (MDRE).
\end{itemize}
Both of them are based on distribution differences of semantic representations between normal examples and adversarial examples.
They achieve state-of-the-art results across a range of attack methods and domains. 

\section{Related Work}
\label{sec:RelatedWork}
In this section, we briefly review state-of-the-art work on defending neural networks against both image and textual adversarial examples.

\par{\textbf{Image Adversarial Defences:}}  Adversarial training~\citep{goodfellow2014explaining} using adversarial examples to augment training data or adding an adversarial objective to a loss function, and defensive distillation framework \citep{papernot2016distillation} which transfers knowledge between same structured teacher and student models, are two effective proactive defence methods. 
For reactive defences, \citet{feinman2017detecting, ma2018characterizing, papernot2018deep, lee2018simple} have all proposed approaches that use the learned representations of the classifier that the attacker is trying to fool, and then with a variety of techniques to identify characteristics of the adversarial examples' learned representations that permit the detection of whether a data point is adversarial or original; these techniques involve kernel density estimations in a feature space of a last hidden layer and Bayesian uncertainty estimates, Local Intrinsic Dimensionality, Deep k-Nearest Neighbors, and Mahalanobis distance-based confidence scores respectively.

\par{\textbf{Textual Adversarial Defences:}} Adversarial training \citep{goodfellow2014explaining} is a commonly used defence method to augment training data with adversarial examples and their correct labels,  
which has been effective in 
\citet{li-etal:2016:ACL}, \citet{li-etal:2017:EACL}, \citet{ribeiro-etal-2018-semantically}, and \citet{ebrahimi-etal-2018-hotflip}, but has limited utility in \citet{pruthi-etal-2019-combating} and \citet{jia-liang-2017-adversarial}.
\citet{jia-etal-2019-certified} applies interval bound propagation (IBP) to minimize an upper bound of possible candidate sentences' losses when facing word substitution adversaries.
\citet{jones-etal-2020-robust} introduced robust encodings (RobEn) to cluster words and typos, and produced one encoding for each cluster to harness adversarial typos.
\citet{zhou-etal-2019-learning} proposed the learning to discriminate perturbations (DISP) framework to block character-level and word-level adversarial perturbations by recognising and replacing perturbed words. 
\citet{mozes-etal-2021-frequency} noticed and verified a characteristic of word-level adversaries that replacement words are less likely to occur than their substitutions, therefore, they constructed a rule-based, model-agnostic frequency-guided word substitutions (FGWS) algorithm, which is the only existing textual reactive defence method as far as we know.

\section{Methods}
\label{sec:method}

\begin{algorithm*}[ht]
    \caption{\small{MultiDistance Representation Ensemble Method (MDRE)}}
    \label{al:mdre}
    \textbf{Input:} \\
    \hspace*{\algorithmicindent} $\D = \{\boldsymbol{X}^{(train)}, \boldsymbol{X}^{(norm)}, \boldsymbol{X}^{(adv)}\}$: a dataset; there are $k$ examples in $\boldsymbol{X}^{(norm)}$ and $\boldsymbol{X}^{(adv)}$\\
    \hspace*{\algorithmicindent} $H$: an array containing $m$ representation learning models \\
    \hspace*{\algorithmicindent} $g: \R^m \rightarrow \{0, 1\}$: a binary classification model (MDRE) \\
    \hspace*{\algorithmicindent} $f: \R^n \rightarrow \R^l$: a deep neural net that is the target model for an adversarial attack \\
    \textbf{Output:} \\
    \hspace*{\algorithmicindent} Detection accuracy of MDRE: $acc$
    \begin{algorithmic}[1]
    \State Initializing inputs and labels of $g$: $\boldsymbol{X}=zeros[2k, m]$, $\boldsymbol{y}=zeros[2k]$
    \State Computing examples' predictions from $f$ of $\D$: $\{\boldsymbol{\hat{y}}^{(train)}, \boldsymbol{\hat{y}}^{(norm)}, \boldsymbol{\hat{y}}^{(adv)}\}$
    \For{$j \in \{0, \cdots, m-1\}$}
    \State Computing examples' representations from $H[j]$  of $\D$: $\{\boldsymbol{V}^{(train)}_j, \boldsymbol{V}^{(norm)}_j, \boldsymbol{V}^{(adv)}_j\}$
    \For{$i \in \{0, \cdots, k-1\}$}
    \State Calculating $d^{(norm)}_j, d^{(adv)}_j$ for examples $\boldsymbol{X}^{(norm)}_i, \boldsymbol{X}^{(adv)}_i$
    \State $\boldsymbol{X}[i, j] = d^{(norm)}_j$, $\boldsymbol{y}[i] = 0$
    \State $\boldsymbol{X}[k+i, j] = d^{(adv)}_j$, $\boldsymbol{y}[k+i] = 1$
    \EndFor
    \EndFor
    \State Training $g$ by randomly choosing $80\%$ of $\{(\boldsymbol{X}_{i,:}, \boldsymbol{y}_i)\}_{i=0}^{2k-1}$
    \State $acc =$ test accuracy of $g$ using the rest $20\%$ of $\{(\boldsymbol{X}_{i,:}, \boldsymbol{y}_i)\}_{i=0}^{2k-1}$
    \end{algorithmic}
\end{algorithm*}

\subsection{Adapted Local Intrinsic Dimensionality (LID)}
\label{sec:lid}
From among the reactive image processing methods, we selected the Local Intrinsic Dimensionality (LID) approach of \citet{ma2018characterizing} as one that can be directly adapted to textual representations.  
The approach of \citet{ma2018characterizing} uses LID to reveal the local distance distribution for a reference point representation to its neighbours, and uses outputs of each layer from the target deep neural network as an input point representations.
LID was initially presented for dimension reduction \citep{Houle2012Generalized}.
\citet{ma2018characterizing} introduced LID to characterize the local data submanifolds in the vicinity of reference points and detect adversarial samples from their originals.
The LID definition is as follows.

\begin{definition}[Local Intrinsic Dimensionality \citep{ma2018characterizing}]
Given a data sample $x \in X$, let $R > 0$ be a random variable denoting the distance from $x$ to other data samples.
If the cumulative distibution function $F(r)$ of $R$ is positive and continuously differentiable at distance $r > 0$, the LID of $x$ at distance $r$ is given by:
\begin{equation}
    \small
	LID_F(r) \triangleq \lim_{\epsilon\to0} \frac{ln(F((1+\epsilon)\cdot r)/F(r))}{ln(1+\epsilon)} = \frac{r\cdot F'(r)}{F(r)}
\end{equation}
whenever the limit exists.
\end{definition}

To simplify computation, given a reference sample $x \sim \mathcal{P}$, where $\mathcal{P}$ represents the data distribution, the Maximum Likelihood Estimator of the LID at $x$ is defined as follows \citep{ma2018characterizing}:
\begin{equation}\label{eq:lid}
	\widehat{LID}(x) = - \left( \frac{1}{k} \sum_{i=1}^{k} \log \frac{r_i(x)}{r_k(x)} \right) ^{-1}
\end{equation}
where $r_i(x)$ is the distance between $x$ and its $i$th nearest neighbor within a sample of points drawn from $\mathcal{P}$, $k$ is the number of nearest neighbors.
Since the logarithmic function $f(x) = \log_a (x)$ for any base $a$ and the negative reciprocal function $f(x) = -x^{-1}$ are monotonically increasing functions when their independent variables are positive, if neighbors of a reference sample $x$ are compact, its estimated LID from Equation~(\ref{eq:lid}) is smaller, otherwise, its estimated LID is bigger.

When building a binary classifier to detect adversarial examples using LID in \citet{ma2018characterizing}, the inputs are lists of estimated LID from the Equation~(\ref{eq:lid}) of different layers' outputs from the target deep neural net, and adversarial and normal examples are two categories of the classifier.

To adapt this to textual representations,
we implement same technique --- a detection classifier based on LID characterizations derived from different layers' outputs of a deep neural net --- but apply this to a Transformer.  Here we use BERT\textsubscript{BASE} model \citep{devlin-etal-2019-bert}, although in principle any would be suitable.
The $x$ in the Equation (\ref{eq:lid}) is a representation of an input text from a layer's hidden state of the first token of the target (BERT\textsubscript{BASE}) model, since self-attention layers are essential modules of a transformer, and the last layer hidden state of the first token is typically used as a component to build a pooled output, a text representation for a classifier.
Therefore, an input of a detection classifier for an example is a 12-dimensional vector, where each element illustrates the corresponding layer's estimated LID from the BERT\textsubscript{BASE} model.

\subsection{MultiDistance Representation Ensemble Method (MDRE)} 
\label{sec:mdre}

Adversarial examples are constructed by adding imperceptible non-random perturbations to inputs of correctly classified test examples to fool highly expressive deep neural nets into incorrect classifications \citep{szegedy2013intriguing}.
Motivated by the reasoning behind LID expressed in Equation (\ref{eq:lid}), by \citet{feinman2017detecting}'s intuition that adversarial samples lie off the true data manifold, and by \cite{lee2018simple}'s recognition that they are out-of-distribution samples by a class-conditional distribution, we assume that samples with a same predicted label from a deep neural net lie on a data submanifold; an adversarial example is generated because perturbations cause a correctly predicted example to transfer from one data submanifold to another, making it an out-of-distribution sample relative to  training examples from its data submanifold.  
Consequently, we posit that it is likely that the Euclidean distance between an adversarial example $x'$ and the nearest neighbor of $x'$ among training examples with the same predicted label 
as $x'$ is bigger than the Euclidean distance between its corresponding original normal test example $x$ and $x$'s nearest neighbor among training examples with the same predicted label as $x$.

In natural language processing, most inputs of deep neural networks are learned representations by representation learning models nowadays. Even though current methods of representation learning are effective in various tasks \citep{devlin-etal-2019-bert, liu2019roberta, yang2019xlnet, lewis-etal-2020-bart}, semantic meanings and semantic differences between texts from humans' perspective are not perfectly captured by textual representation vectors \citep{liu2020representation}.
In addition, as mentioned in Section \ref{sec:Introduction}, most textual adversarial generation algorithms do not modify representations, which are input feature vectors of deep neural networks, but modify original texts.
Therefore, the assumed characteristic of adversaries in the last paragraph that the Euclidean distances between adversarial examples and their nearest neighbors among training examples in the same submanifolds are bigger than normal examples, may lose efficiency in textual adversarial detection scenarios.
To build a stronger reactive classifier, we use ensemble learning to combine distances between representations learned from multiple representation learning models.
We construct a more effective MultiDistance Representation Ensemble Method (MDRE), as illustrated in Algorithm \ref{al:mdre}.

The MDRE is a supervised binary classification model $g:\R^m \rightarrow \{0, 1\}$.
$m$ is the number of representation learning models; $g$ can be any binary classification model, such as logistic regressions or deep neural nets; $\{0, 1\}$ is the output label set, with $1$ corresponding to adversarial examples, $0$ to normal examples.

The input of MDRE is a matrix $\boldsymbol{X}$ and each row vector of $\boldsymbol{X}$ is $\boldsymbol{X}_{i,:} = (d_0, d_1\cdots, d_{m-1}) \in \R^m$.
The element of this vector $d_j, 0 \leq j \leq m-1$ is a Euclidean distance between a semantic representation of a normal or adversarial example $\boldsymbol{v}$ and a representation of its nearest neighbour among training examples with the same predicted label as $\boldsymbol{v}$ through the $j$-th representation learning model $H[j]$, as $d^{(norm)}_j$ or $d^{(adv)}_j$ in Algorithm~\ref{al:mdre}.
To find a nearest neighbour, we compare Euclidean distances between $\boldsymbol{v}$ and all representations among training examples with the same predicted label as $\boldsymbol{v}$ through $H[j]$.
In Algorithm \ref{al:mdre}, $\boldsymbol{X}^{(norm)}$ consists of normal test examples corresponding to the elements of $\boldsymbol{X}^{(adv)}$, where the elements of $\boldsymbol{X}^{(norm)}$ have correct predictions from the target model $f$, but $\boldsymbol{X}^{(adv)}$ elements have incorrect predictions from $f$.
The training and testing process of MDRE is same as the process of the selected model $g$.

\section{Evaluation}
In this section, we evaluate the utilities of the adapted LID and MDRE by using character-level, word-level, and phrase-level upstream attacks on sentiment analysis and natural language inference tasks, and comparing against several baselines: a language model, DISP \citep{zhou-etal-2019-learning}, and FGWS \citep{mozes-etal-2021-frequency}.
The experimental results demonstrate that the adapted LID and MDRE outperforms these methods on sentiment analysis and natural language inference tasks for word-level and phrase-level attacks.

\subsection{Experimental Setup}
\subsubsection{Tasks}
We apply our approaches and baselines to sentiment analysis and natural language inference tasks, since they are two most commonly used datasets in textual adversarial example generation.
The sentiment analysis task has been the most widely used testbed for generating textual adversarial examples \citep{pruthi-etal-2019-combating, alzantot-etal-2018-generating, ribeiro-etal-2018-semantically,ren-etal-2019-generating,iyyer-etal-2018-adversarial, yoo-qi-2021-towards-improving, li-etal-2020-bert-attack}, making this the natural domain for these experiments; they have also been popularly applied to the natural language inference task \citep{alzantot-etal-2018-generating, iyyer-etal-2018-adversarial, yoo-qi-2021-towards-improving, li-etal-2020-bert-attack, li-etal-2021-contextualized, jin2020bert}, 
so we choose this to explore the generality of our methods.

We use the IMDB dataset \citep{maas-etal-2011-learning} in the sentiment analysis task, which contains 50,000 movie reviews, divided into 25,000 training examples and 25,000 test examples, labelled for positive or negative sentiment.
The average number of words per review in the IMDB dataset is 262 when using the Natural Language Toolkit (NLTK) \citep{bird2009NLTK} to tokenize examples.
We set a maximum sequence length of the IMDB dataset to 512 for all following models.

To test the robustness of our methods, the Multi-Genre NLI (MultiNLI) corpus  \citep{williams-etal-2018-broad} and its mismatched test examples, which are derived from sources that differ from the training examples, are used in the natural language inference task.
The MultiNLI dataset includes 392,702 training examples and 9,832 mismatched testing examples in which global\_label fields are not "-", with three classes: entailment, neutral, and contradiction.
The average and maximum word numbers of the MultiNLI dataset are 34 and 416 respectively, using NLTK word tokenizer.
We set the maximum sequence length for this dataset to 256.

\begin{table*}[ht!]
    \centering
    \small
    \setlength\tabcolsep{1.5pt}
    \begin{tabular}{cccccccc}
    \toprule
    \multirow{2}{4em}{\centering \textbf{Dataset}} & \multirow{2}{4em}{\centering \textbf{Training.}} & \multirow{2}{5em}{\centering \textbf{Validation.}} &\multirow{2}{4em}{\centering \textbf{Testing.}} & \multirow{2}{9em}{\centering \textbf{Correctly Predicted Test Examples}} & \multicolumn{3}{c}{\textbf{Adversarial/Original Examples}} \\ [0.5ex]
    & & & & &  \multicolumn{1}{c}{character-level} & \multicolumn{1}{c}{word-level} & \multicolumn{1}{c}{phrase-level}  \\
    \midrule
    IMDB & 20,000 & 5,000 & 25,000 & 23,226 & 12,299 & 9,627 & 6,315  \\
    MultiNLI & 314,162 & 78,540 & 9,832 & 8,062 & 7,028 & 3,240 & 4,340 \\
    \bottomrule
    \end{tabular}
    \caption{\small{The number of examples used in experiments}}
    \label{tab:num_examples}
    \vspace{-12pt}
\end{table*}

\subsubsection{Attack Methods}
We implement three widely used attack methods using character-level, word-level, and phrase-level perturbations to construct adversarial examples.
For all types of attacks, we take the BERT\textsubscript{BASE} model as the target model, indicating that adversaries have different predictions with their originals by the BERT\textsubscript{BASE} model. 

\par{\textbf{Character-level.}}
The character-level attack is from \citet{pruthi-etal-2019-combating}, which applies swapping, dropping, adding, and keyboard mistakes to a randomly selected word of an original example.

\begin{itemize}[noitemsep]
	\item Swapping: swapping two adjacent internal characters.
	\item Dropping: removing an internal character.
	\item Adding: internally inserting a new character.
	\item Keyboard mistakes: substituting an internal character with one of its adjacent characters in keyboards.
\end{itemize}

Here, we set maximum numbers of perturbations to half of the maximum sequence lengths of datasets; consequently, for the IMDB dataset, the maximum number of attacks is 256, and for the MultiNLI dataset is 128.
If after achieving this number, the prediction of the perturbed text is still consistent with the original example, these attacks fail, and no character-level adversarial example constructed for this original example.

\par{\textbf{Word-level.}}
We use a method from~\citet{alzantot-etal-2018-generating}, which is an effective and widely cited word-level threat method.
Their approach randomly selects a word in a sentence, replaces it with its synonymous and context fitted word according to the GloVe word vectors \citep{pennington-etal-2014-glove}, counter-fitting word vectors \citep{mrksic-etal-2016-counter}, and the Google 1 billion words language model \citep{chelba2013one}, and applies population-based genetic algorithms from the natural selection using a combination of crossover and mutation to generate next adversarial generations. 

While effective, the initial algorithm is somewhat inefficient and computationally expensive.
In implementing this method, \citet{jia-etal-2019-certified} found that computing scores from the Google 1 billion words language model \citep{chelba2013one} for each iteration in this approach causes its inefficiency; to improve this, they used a faster language model and  prevented semantic drift, which is synonyms picked from previous iterations also apply the language model to select words from their neighbour lists.
In our experiments, we adapt these modifications by using a faster Transformer-XL architecture~\citep{dai-etal-2019-transformer} pretrained on the WikiText-103 dataset \citep{merity2016pointer}, and not allowing the semantic drift, so that we compute all test examples words' neighbours before attacks.

In this attack, we also set maximum numbers of perturbations, which are one fifth of the maximum sequence lengths; therefore, for the IMDB dataset is 102, and for the MultiNLI dataset is 51.
For an original test example, if the number of attacks reaches this threshold but predictions do not change, no corresponding adversarial example is constructed for this original example.

\par{\textbf{Phrase-level.}}
The phrase-level attack is from \citet{ribeiro-etal-2018-semantically}, which uses translators and back translators to generate adversarial examples. 
As far as we know, this is the only phrase-level perturbation technique that can be used for paragraph-length text.
Their approach --- termed semantically equivalent adversaries (SEAs) --- translates an original sentences into multiple pivot languages, then translates them back to the source language.
If there is a back translated sentences that is semantically equivalent to the original sentences, measured by a semantic score greater than a threshold, and it has a different prediction with the original sentences, then it is an adversarial example.
Otherwise, this original example has no relevant adversaries.

\subsubsection{Target Model}

The BERT\textsubscript{BASE} model is implemented as a target model for these three attacks, by which adversarial examples are misclassified.
We apportion training sets on both datasets into training subsets and validation subsets, with an 80-20 split.
After training, the models achieve $92.90\%$ test accuracy on the IMDB dataset, and for the MultiNLI mismatched test set is $82.01\%$.
The correctly predicted test examples are preserved for subsequent attack processes.
After attacks, adversarial examples and their corresponding normal test examples maintain for following detectors as negative and positive examples;
in this, we follow the experimental setup used for evaluating reactive defences in the image processing literature \citep{ma2018characterizing} with an 80/20 training/test split.
The number of examples used on the IMDB and MultiNLI datasets and number of originals and adversaries after attacks are shown in Table \ref{tab:num_examples}.

\begin{table*}[ht!]
    \centering
    \small
    \setlength\tabcolsep{1.5pt}
    \captionsetup{justification=centering}
    \begin{tabular}{cccccc}
    \toprule
    Dataset & Attack Method & BERT\textsubscript{BASE} & RoBERTa\textsubscript{BASE} & XLNet\textsubscript{BASE} & BART\textsubscript{BASE} \\ [0.5ex]
    \midrule
    \multirow{3}{3em}{IMDB} & Character-level & 0.3656  & 0.8613  & 0.5770  & 0.8286  \\
    & Word-level &  0.6999 &  0.8714 & 0.7918  &   0.8425 \\
    & Phrase-level & 0.1827 & 0.3224  & 0.3289  & 0.3010  \\
    \midrule
    \multirow{3}{4em}{MultiNLI} & Character-level & 0.4848 &  0.7104 & 0.6670  & 0.6457  \\
    & Word-level & 0.6864
  & 0.7068  & 0.6870  &  0.6296  \\
    & Phrase-level & 0.2795
  & 0.3899  & 0.3698  & 0.3325 \\
    \bottomrule
    \end{tabular}
    \caption{\small{The accuracy of adversarial examples}}
    \label{tab:acc_noise_adv}
    \vspace{-12pt}
\end{table*}

\subsubsection{Detection Methods}
\label{sec:detect_methods}
We evaluate three baselines in addition to the adapted LID and MDRE in these experiments.  

\par{\textbf{A language model.}}
The first baseline is built from a 
language model since even though most attack algorithms intend to construct semantically and syntactically similar adversaries, many textual adversaries are abnormal and ungrammatical, as shown in Table \ref{tab:text_adv}.
We use the Transformer-XL~model pretrained on the WikiText-103 dataset from Hugging Face transformers \citep{wolf-etal-2020-transformers}, and obtain language model scores for texts as the product of words prediction proportion scores.
We construct a detection classifier by using a logistic regression model with language model scores as inputs; the model acts to learn a threshold on scores to distinguish adversarial examples.

\par{\textbf{Learning to Discriminate Perturbations (DISP) \citep{zhou-etal-2019-learning}.}}
Our second baseline is the DISP framework, which is the only comparable technique for detecting textual adversarial examples across character-level and word-level attacks to our knowledge.
DISP consists of three components: perturbation discriminator, embedding estimator, and hierarchical navigable small word graphs.
The perturbation discriminator identifies a set of character-level or word-level perturbed tokens;
the embedding estimator predicts embeddings for each perturbed token; then, hierarchical navigable small word graphs map these embeddings to actual words to correct adversarial perturbations.
DISP is not itself designed as a adversarial example detector, but we adapt it for that task: 
if an adversarial example rectified by DISP predicts the same class as the target model predicts for the corresponding initial original example, or the prediction of a normal (non-adversarial) example rectified by DISP isn't changed, we consider DISP to have been successful in its detection.  Otherwise, it is not.
Since DISP is designed for character-level and word-level attacks, we do not apply it to phrase-level attacks.

\par{\textbf{Frequency-guided word substitutions \\ (FGWS)~\citep{mozes-etal-2021-frequency}.}}
Our third baseline is FGWS.
\citet{mozes-etal-2021-frequency} noticed, and verified using hypothesis testing, that a characteristic of word-level adversaries was that replacement words are less likely to occur than their substitutions.
They use this feature to construct a rule-based, model-agnostic frequency-guided word substitutions (FGWS) algorithm which distinguishes adversarial examples by replacing infrequent words with their higher frequency synonyms.
If the replacements cause prediction confidence changes exceeding a threshold, these examples are deemed adversarial examples.
FGWS is only designed to be applied to  word-level attacks.
They use WordNet \citep{Fellbaum2005-FELWAW} and GloVe vectors \citep{pennington-etal-2014-glove} to find neighbors of a word.
A word frequency is its number of occurrences in the corresponding dataset's training examples; infrequent words are defined as those words whose frequencies are lower than a threshold.
They set this threshold to be the frequency of the word at the \{$\mathit{0}$-th, $\mathit{10}$-th, $\cdots$, $\mathit{100}$-th\} percentile of word frequencies in training set.
If the prediction confidence differences between sequences with replaced words and their corresponding original sequences are higher than a threshold, the original sequences are assumed to be adversarial examples.
They set this threshold to the $\mathit{90\%}$-th confidence difference between words substituted validation set and original validation set in their experiment.

\par{\textbf{Adapted Local Intrinsic Dimensionality (LID)}}
Following the characterization of our adapted LID from Section~\ref{sec:lid}, we use the BERT\textsubscript{BASE} model as in the above baselines.
We implement a 
logistic regression model as the detection classifier as \citet{ma2018characterizing}, and the neighborhood size $k$ is tuned using a grid search over 100, 1000, and the range [10, 42) with a step size 2.

\par{\textbf{MultiDistance Representation Ensemble Method (MDRE).}}
In MDRE, we set $m=4$, $H=$ [BERT\textsubscript{BASE}, RoBERTa\textsubscript{BASE}, XLNet\textsubscript{BASE}, BART\textsubscript{BASE} ], and $g$ is a logistic regression model.
See Algorithm \ref{al:mdre} for more information of notations.

\begin{table}[b]
    \centering
    \vspace{-5pt}
    \small
    \setlength\tabcolsep{1.5pt}
    \begin{tabular}{ccccc}
    \toprule
         Dataset & BERT\textsubscript{BASE} & RoBERTa\textsubscript{BASE} & XLNet\textsubscript{BASE} & BART\textsubscript{BASE} \\
         \midrule
         IMDB & 0.9290 & 0.9532 & 0.9336 & 0.9429 \\
         MultiNLI & 0.8201 & 0.8671 & 0.8630 & 0.8455 \\
    \bottomrule
    \end{tabular}
    \caption{\small{The accuracy of normal test examples}}
    \label{tab:acc_normal}
    \vspace{-12pt}
\end{table}

\begin{table*}[ht!]
    \centering
    \small
    \setlength\tabcolsep{1.5pt}
    \begin{tabular}{ccccc}
    \toprule
    Dataset & Detecting Method & Character-level Attack & Word-level Attack & Phrase-level Attack \\ [0.5ex]
    \midrule
    \multirow{5}{*}{IMDB}
    & Language Model & 0.4996 & 0.4966 & 0.4838 \\
    & DISP & 0.8936 & 0.7714 & --- \\
    & FGWS & --- & 0.7958 & --- \\
    & LID & 0.9142 & \textbf{0.8406} & 0.9093 \\
    & MDRE & \textbf{0.9193} & 0.7562 & \textbf{0.9505} \\
    \midrule
    \multirow{5}{*}{MultiNLI}
    & Language Model & 0.4932 &  0.4707 & 0.4997 \\
    & DISP & \textbf{0.7496} & 0.6137 & ---\\
    & FGWS & --- & 0.6128 & --- \\ 
    & LID & 0.7328 & 0.5849 & 0.6146 \\
    & MDRE & 0.7016 & \textbf{0.6319} & \textbf{0.6809} \\
    \bottomrule
    \end{tabular}
    \caption{The accuracy for detection classifiers}
    \label{tab:result}
    \vspace{-12pt}
\end{table*}

\subsection{Experimental Results}
Before discussing the effectiveness of the detection classifiers,
Table \ref{tab:acc_normal} and Table~\ref{tab:acc_noise_adv} show the accuracy of the sentiment analysis and natural language inference classifiers on normal and adversarial examples from four models with three types of attacks. 
The BERT\textsubscript{BASE} model is the target model in terms of generating all kinds of adversaries --- that is, the adversarial examples are specifically designed to defeat the BERT\textsubscript{BASE} model so all adversarial instance predictions are incorrect, therefore, the accuracy is 0.
However, when we use a different random seed which also modify the order of training examples to fine-tune another BERT\textsubscript{BASE} model used for prediction,
its parameters is different from the parameters of the BERT\textsubscript{BASE} model used before.
The accuracy of adversaries slightly increases, indicating that BERT model parameters do not converge but fluctuate when using stochastic or mini-batch gradient descent.

Results for detection method accuracy are in Table~\ref{tab:result}.
Adapted LID and MDRE work better than the baselines, except for DISP against character-level attacks on MultiNLI dataset, where the adapted LID is a close second.
The detection accuracy on the MultiNLI dataset is lower than the IMDB dataset, although this is not a surprise.
It uses the mismatched test set of the MultiNLI dataset which makes the task more challenging.
The results show that the adapted LID and MDRE are sensitive to sample distributions, so if some normal test examples representations are from a different distribution of training samples representations, such as noise examples, they will influence their performance.

Adapted LID is often close to MDRE.  
It is higher in  word-level attack on the IMDB dataset and character-level attack on the MultiNLI dataset, 
but it is lower on phrase-level attacks.
Relative to its initial application on image classification tasks, 
the performance of the adapted LID approach is worse.
Most accuracy of LID on image adversarial attacks on CIFAR-10, CIFAR-100, and SVHN datasets are over or near 90\% \cite{CVPRPaper}.
However, in our experiments, the average accuracy of the adapted LID is about 77\% (against majority class baseline of 50\%).
This reveals the difficulty of detecting textual adversarial examples.

The performance of the language model is similar to random guess, since the ratio between positive (normal) and negative (adversarial) examples is 1:1.
We observed that language model prediction proportion scores are sensitive to the number of words in examples because each word scores is between 0 to 1 and more words leads to lower scores.
In addition, in some contexts, scores for synonyms or typos which are out-of-dictionary words, are lower but close to scores of original words, which do not have the large differences that might be expected.

DISP effectively applies the bidirectional language model feature of the BERT\textsubscript{BASE} model and builds a powerful perturbation discriminator, which labels character-level or word-level perturbed tokens to 1, and unperturbed tokens to 0.
The perturbation discriminator achieves $F_1$ scores of 95.06\% on the IMDB dataset and 97.67\% on the MultiNLI dataset, using their own adversaral attack methods. 
However, the embedding estimator predicts embeddings through inputting 5-grams with masked middle tokens to a BERT\textsubscript{BASE} model with one layer feed-forward head on top and outputting embeddings of these masked tokens from 300-dimensional pretrained FastText English word vectors \citep{mikolov-etal-2018-advances}.
This is challenging and restricts the overall performance of DISP.

Intuitively, adversaries' predictions are different from their original counterparts, which are ordinary language; therefore, adversaries may contain rare and infrequent words.
According to an English word frequency dataset,\footnote{The english word frequency: \url{https://www.kaggle.com/rtatman/english-word-frequency}} some words frequencies in examples of \citet{alzantot-etal-2018-generating} are shown in Table \ref{tab:word_freq}.
\begin{table}[h]
    \centering
    \small
    \begin{tabular}{p{0.08\textwidth}p{0.1\textwidth}p{0.08\textwidth}p{0.08\textwidth}}
    \toprule
     \textbf{org.}  & \textbf{org. freq.} & \textbf{sub.} & \textbf{sub. freq.} \\\midrule
     \multirow{2}{*}{terrible} & \multirow{2}{*}{{\small 8,610,277}} & horrific & {\small 1,017,211} \\\cline{3-4}
     & & horrifying & {\small 491,916} \\
     considered & {\small 57,378,298} & regarded & {\small 6,892,622} \\
     kids & {\small 96,602,880} & youngstars & --- \\
     runner & {\small 7,381,022} & racer & {\small 3,625,077} \\
     battling & {\small 1,340,424} & --- & --- \\
     strives & {\small 1,415,683} & --- & --- \\\bottomrule
    \end{tabular}
    \caption{\small Original and modified sample words frequencies in examples of \citet{alzantot-etal-2018-generating}}
    \label{tab:word_freq}
    \vspace{-12pt}
\end{table}
We can find that the intuition is correct that replacement words frequencies drop compared with substitutions; however, they may be higher than other normal words.
Therefore, using one threshold makes it difficult to separate adversarially substituted words from all normal words.
Alternative approaches to applying the characteristic of adversarial words frequencies may work better.  
We note that it is perhaps surprising, then, that our representation-based detection methods outperform FGWS that do incorporate frequency information from the raw text input.  This underscores the usefulness of the distributional information available in the learned representations.

We show detection methods applied to examples from the MultiNLI dataset in the Appendix~\ref{sec:app-results} supplement.

\begin{table*}[ht!]
    \centering
    \small
    \setlength\tabcolsep{1.5pt}
    \begin{tabular}{ccccc}
    \toprule
    Dataset & Detecting Method & Character-level Attack & Word-level Attack & Phrase-level Attack \\ [0.5ex]
    \midrule
    \multirow{3}{*}{IMDB} 
    & MDRE\textsubscript{BERT} & 0.8941 & 0.7541 & 0.9129 \\
    & MDRE\textsubscript{RoBERTa} & 0.8606 & 0.6645 & 0.9287 \\
    & MDRE\textsubscript{XLNet} & 0.7226 & 0.5962 & 0.7819 \\
    & MDRE\textsubscript{BART} & 0.8951 & 0.6858 & 0.9327 \\
    \midrule
    \multirow{3}{*}{MultiNLI}
    & MDRE\textsubscript{BERT} & 0.6102 & 0.5903 & 0.6382  \\
    & MDRE\textsubscript{RoBERTa} & 0.6853 & 0.5903 & 0.6526 \\
    & MDRE\textsubscript{XLNet} & 0.6323 & 0.6227 & 0.6452 \\
    & MDRE\textsubscript{BART} & 0.6824 & 0.6366 &  0.6740 \\
    \bottomrule
    \end{tabular}
    \caption{The accuracy of detection classifiers for ablation analysis of MDRE}
    \label{tab:ablation}
    \vspace{-12pt}
\end{table*}
\subsection{Ablation Analysis of MDRE}
The key ideas behind MDRE is that (1) adversarial examples are out-of-distribution samples relative to training examples from their data submanifolds and (2) ensemble learning can help identify this.
Therefore, we combine four representation learning models: BERT\textsubscript{BASE}, RoBERTa\textsubscript{BASE}, XLNet\textsubscript{BASE}, and BART\textsubscript{BASE} to produce MDRE as described in Section \ref{sec:detect_methods}.
In order to explore the effects of these two components and each representation learning model, we apply MDRE\textsubscript{BERT}, MDRE\textsubscript{RoBERTa}, MDRE\textsubscript{XLNet}, MDRE\textsubscript{BART} models, where $m=1$, $H=$ [BERT\textsubscript{BASE}], [RoBERTa\textsubscript{BASE}], [XLNet\textsubscript{BASE}], and [BART\textsubscript{BASE}] respectively.

The results are shown in Table~\ref{tab:ablation} which reveals all models work in detecting textual adversarial examples: the detection accuracy on both the IMDB and MultiNLI datasets, and all upstream adversarial attacks is substantially higher than random guess (50\%).
Comparing with the results of MDRE on the IMDB and MultiNLI datasets from Table \ref{tab:result}, 
ensemble learning helps to build a stronger detector except word-level attack on the MultiNLI dataset.

\section{Conclusion and Future work}
In this paper, we adapted Local Intrinsic Dimensionality (LID) method \citep{ma2018characterizing} from image processing  and proposed a simple and general textual adversarial reactive detector, MultiDistance Representation Ensemble Method (MDRE), based on the distribution characteristics of adversarial examples representations, that they are out-of-distribution samples and lie off the true data manifold.
The experimental results show adapted LID and MDRE achieve state-of-the-art results on detecting character-level, word-level, and phrase-level adversaries on the IMDB dataset as well as on the later two with respect to the MultiNLI dataset. The results show that it is possible to construct adversarial example detectors using only the learned representations, and not relying on various textual substitution processes as in the baselines.

As discussed in Section \ref{sec:method}, adapted LID uses estimated Local Intrinsic Dimensionality on text representations form different layers outputs of a target model, and MDRE is implemented on Euclidean distances between samples' representations and representations of their nearest neighbors among the training examples with the same predicted labels from different representation learning models, to characterise representation distribution differences between adversarial examples and normal examples.
In terms of future work and the LID approach, \citet{athalye-etal:2018:ICML} found that in the image processing space, LID is vulnerable to their Backward Pass Differentiable Approximiation (BPDA) attack; it would be useful to investigate whether this is the case in the text space, and if so, other detection methods from image processing may be worth looking into.
With respect to MDRE, as it is a kind of nearest-neighbour ensembling approach, looking into other possibilities falling within that space could be productive.
More generally, exploring more effective distribution characteristics of data semantic representations among adversarial and normal examples, may help to build better detectors.

\setlength{\bibsep}{0pt}
\bibliography{anthology,custom_short}
\bibliographystyle{acl_natbib}

\newpage
\appendix

\section{Experimental Results Samples}
\label{sec:app-results}

Samples of outputs produced by the word-level attack and four detection classifiers on the MultiNLI dataset are shown in Table \ref{tab:sample}, to illustrate where some detection methods work while others do not.  
The DISP and FGWS baselines both also produce `corrected' text; their outputs are included here.

In our experiments, the best accuracy of FGWS is when the frequency threshold is 92 and the threshold for the difference in prediction confidence is about 0.1916, therefore, if a word appears in the MultiNLI dataset training set and its occurrence frequency in the training corpus is lower than 92, it will be replaced by another word that is semantically similar and has higher occurrence frequency in the training set. 
If after transformations, the difference in prediction confidence before and after exceeds 0.1916, this example is considered as an adversarial example.

In example (a), MDRE, adapted LID, and DISP are successful, but FGWS does not detect this word-level adversarial example, because the occurrence frequency of the substituted word \textit{shopping} for \textit{store} is 1153 which is higher than the threshold 92, but original words \textit{mentioning} and \textit{buffer} are replaced by \textit{name} and \textit{pilot} respectively, since their occurrence frequencies are 67 and 30 in the MultiNLI training set which are lower than the threshold 92. 
From the DISP output, we can see that it detects \textit{shopping} as a problem word and it is substituted by \textit{do}.

\begin{table*}[h!]
    \begin{subtable}[h]{\textwidth}
        \centering
        \small
        \begin{supertabular}{p{\textwidth}}
        \toprule
        \textbf{Original example prediction: Entailment}\\\midrule
        \textbf{Premise}: Finally, it might be worth mentioning that the program has the capacity to store in a temporary memory buffer about 100 words (proper names, for instance) that it has identified as not stored in its dictionary. \\
        \textbf{Hypothesis}: It's possible to \textbf{\textcolor{darkspringgreen}{store}} words in a temporary dictionary, if they don't appear in a regular dictionary. 
        \\\midrule
        \textbf{Word-level adversarial example prediction: Neutral}
        \\\midrule
        \textbf{Premise}: Finally, it might be worth \textbf{\textcolor{orange}{mentioning}} that the program has the capacity to store in a temporary memory \textbf{\textcolor{orange}{buffer}} about 100 words (proper names, for instance) that it has identified as not stored in its dictionary.\\
        \textbf{Hypothesis}: It's possible to \textbf{\textcolor{orange}{shopping}} words in a temporary dictionary, if they don't appear in a regular dictionary.
         \\\midrule
         \textbf{DISP output of this word-level adversarial example}
         \\\midrule
        \textbf{Premise}: Finally, it might be worth \textbf{\textcolor{darkspringgreen}{that}} that the program has the capacity to store in a temporary memory buffer about 100 words (proper names, for instance) that it has identified as not stored in its dictionary. \\
        \textbf{Hypothesis}: It's possible to \textbf{\textcolor{darkspringgreen}{do}} words in a temporary dictionary, if they don't appear in a regular dictionary.
         \\\midrule
         \textbf{FGWS output of this word-level adversarial example}
         \\\midrule
        \textbf{Premise}: Finally, it might be worth \textbf{\textcolor{darkspringgreen}{name}} that the program has the capacity to store in a temporary memory \textbf{\textcolor{darkspringgreen}{pilot}} about 100 words (proper names, for instance) that it has identified as not stored in its dictionary. \\
        \textbf{Hypothesis}: It's possible to shopping words in a temporary dictionary, if they don't appear in a regular dictionary. \\
        \bottomrule
        \end{supertabular}
        \caption{\small{An example with MDRE, adapted LID, and DISP correct predictions; \\ FGWS and the language model incorrect predictions on the adversarial example}}
    \end{subtable}
    \begin{subtable}[h]{\textwidth}
        \small
        \begin{supertabular}{p{\textwidth}}
            \toprule
            \textbf{Original example prediction: Neutral}\\\midrule
            \textbf{Premise}: I've been going up as a progress in \textbf{\textcolor{darkspringgreen}{school}}, so I, it will be a good \textbf{\textcolor{darkspringgreen}{change}} for me. \\
            \textbf{Hypothesis}: I \textbf{\textcolor{darkspringgreen}{think}} further \textbf{\textcolor{darkspringgreen}{change}} can \textbf{\textcolor{darkspringgreen}{help}} me \textbf{\textcolor{darkspringgreen}{improve}} even more.
            \\\midrule
            \textbf{Word-level Adversarial Example prediction: Entailment}
            \\\midrule
            \textbf{Premise}: I've been going up as a progress in \textbf{\textcolor{orange}{teaching}}, so I, it will be a good \textbf{\textcolor{orange}{amendment}} for me . \\
            \textbf{Hypothesis}: I \textbf{\textcolor{orange}{thought}} further \textbf{\textcolor{orange}{alter}} can \textbf{\textcolor{orange}{support me}} \textbf{\textcolor{orange}{improvement}} even more.
             \\\midrule
             \textbf{DISP output of this word-level adversarial example}
             \\\midrule
            \textbf{Premise}: I've been going up as a progress in teaching, so I \textbf{\textcolor{darkspringgreen}{think}} it will be a good amendment for me. \\
            \textbf{Hypothesis}: I thought further \textbf{\textcolor{darkspringgreen}{that}} can support \textbf{\textcolor{darkspringgreen}{and}} improvement even more.
             \\\midrule
             \textbf{FGWS output of this word-level adversarial example}
             \\\midrule
            \textbf{Premise}: I've been going up as a progress in teaching, so I, it will be a good amendment for me. \\
            \textbf{Hypothesis}: I thought further alter can support me improvement even more.
             \\\bottomrule
        \end{supertabular}
        \caption{\small{An example with adapted LID correct prediction; \\ MDRE, DISP, FGWS, and the language model incorrect predictions on the adversarial example}} 
    \end{subtable}
    \begin{subtable}[h]{\textwidth}
        \small
        \begin{supertabular}{{p{\textwidth}}}
        \toprule
        \textbf{Original example prediction: Contradiction}\\\midrule
        \textbf{Premise}: Increased profit came from missing fewer sales by being in stock a higher percentage of the time.\\
        \textbf{Hypothesis}: Profits \textbf{\textcolor{darkspringgreen}{declined}} because less sales were missed.
        \\\midrule
        \textbf{Word-level adversarial example prediction: Entailment}
        \\\midrule
        \textbf{Premise}: Increased profit came from missing fewer sales by being in stock a higher percentage of the time .\\
        \textbf{Hypothesis}: Profits \textbf{\textcolor{orange}{dipped}} because less sales were missed .
         \\\midrule
         \textbf{DISP output of this word-level adversarial example}
         \\\midrule
        \textbf{Premise}: Increased profit came from missing fewer sales by being in stock a higher percentage of the time . \\
        \textbf{Hypothesis}: Profits dipped because less sales were missed .
         \\\midrule
         \textbf{FGWS output of this word-level adversarial example}
         \\\midrule
        \textbf{Premise}: Increased profit came from missing fewer sales by being in stock a higher percentage of the time . \\
        \textbf{Hypothesis}: Profits \textbf{\textcolor{darkspringgreen}{duck}} because less sales were missed . 
        \\\bottomrule
        \end{supertabular}
        \caption{\small{An example with MDRE, FGWS correct predictions; \\ adapted LID, DISP, and the language model incorrect predictions on the adversarial example}} 
    \end{subtable}
    \caption{\small{Examples of detection results on the MultiNLI dataset}}
    \label{tab:sample}
    \vspace{-8pt}
\end{table*}

In example (b), only adapted LID is successful.  
This is an odd (but not atypical) example in that the premise is not grammatical in written English, which might cause its representation differ from normal examples and lead MDRE to predict wrong. 
However, the prediction confidence about the premise and the hypothesis are unrelated from BERT\textsubscript{BASE} model is 90.49\%, therefore, the word-level adversarial method have to make many changes to both premise and hypothesis to fool the target classifier.
All words occurrence frequencies are above the threshold 92.
FGWS and DISP fail in detecting most substitution words in this adversarial example.

In example (c), only MDRE and FGWS are successful.  As with example (a), there is only a single word change.  
Even though \textit{dipped} is not an infrequent word, there are only 45 occurrences in the MultiNLI training corpus, which is lower than the threshold 92, so FGWS detects it.
The language model detector doesn't detect these three adversarial examples, since it fails to learn a threshold on the language model scores to separate normal and adversarial examples, and predict nearly all examples as normal examples.
\end{document}